\documentclass[fleqn,10pt]{wlscirep}
\usepackage[utf8]{inputenc}
\usepackage[T1]{fontenc}
\title{Automated Human Cell Classification in Sparse Datasets using Few-Shot Learning}

\author[1,*]{Reece Walsh}
\author[1]{Mohamed H. Abdelpakey}
\author[1]{Mohamed S. Shehata}
\author[2]{Mostafa M. Mohamed}
\affil[1]{Department of Computer Science, Mathematics, Physics and Statistics, University of British Columbia, Kelowna, Canada}
\affil[2]{Helwan University, Department of Computer Science, Egypt}

\affil[*]{reece.walsh@ubc.ca}


\begin{abstract}
Classifying and analyzing human cells is a lengthy procedure, often involving a trained professional. In an attempt to expedite this process, an active area of research involves automating cell classification through use of deep learning-based techniques. In practice, a large amount of data is required to accurately train these deep learning models. However, due to the sparse human cell datasets currently available, the performance of these models is typically low. This study investigates the feasibility of using few-shot learning-based techniques to mitigate the data requirements for accurate training. The study is comprised of three parts: First, current state-of-the-art few-shot learning techniques are evaluated on human cell classification. The selected techniques are trained on a non-medical dataset and then tested on two out-of-domain, human cell datasets. The results indicate that, overall, the test accuracy of state-of-the-art techniques decreased by at least 30\% when transitioning from a non-medical dataset to a medical dataset. Reptile and EPNet were the top performing techniques tested on the BCCD dataset and HEp-2 dataset respectively. Second, this study evaluates the potential benefits, if any, to varying the backbone architecture and training schemes in current state-of-the-art few-shot learning techniques when used in human cell classification. To this end, the best technique identified in the first part of this study, EPNet, is used for experimentation. In particular, the study used 6 different network backbones, 5 data augmentation methodologies, and 2 model training schemes. Even with these additions, the overall test accuracy of EPNet decreased from 88.66\% on non-medical datasets to 44.13\% at best on the medical datasets. Third, this study presents future directions for using few-shot learning in human cell classification. In general, few-shot learning in its current state performs poorly on human cell classification. The study proves that attempts to modify existing network architectures are not effective and concludes that future research effort should be focused on improving robustness towards out-of-domain testing using optimization-based or self-supervised few-shot learning techniques.

\end{abstract}
\begin{document}

\flushbottom
\maketitle

\thispagestyle{empty}

\section*{Introduction}

Visual analysis of human cells has long served as a steadfast diagnostic tool for a variety of potential ailments. Examples of these procedures include blood smear tests used for diagnosis of blood conditions or skin biopsies used for discovery of epidermal diseases. Analysis of human cells, however, can be a time consuming task, requiring the attention of a trained professional for significant portions of time. Automated cell counters, or machines of a similar likeness, have alleviated some of the less complex, monotonous tasks. Automated classification of complex cell structures, though, remains a difficult goal due to large variations in cell shape, differing cell-image capturing methodologies, and variance in cell staining protocols. Experienced professionals, however, are capable of overcoming these obstacles and identifying a wide variety of human cell types in adverse visual conditions. This raises the question if similar flexible understanding of cells can be instilled within a given model.

Work within the field of artificial intelligence (AI) has historically struggled to achieve performance similar to human perception. Some of the earliest work, such as Strachey's Draughts program \cite{link2012programming}, pushed the capabilities of technology at the time while attempting to employ a learning mechanism to teach the machine about a given task. Over the years, Reasoning-as-Search \cite{mccorduck2004machines}, Expert Systems \cite{jackson1998introduction} and other techniques have been proposed as paradigms for enabling intelligent processing in computer programs. Today, however, Backpropagation-based techniques \cite{rumelhart1986learning} are largely favoured, with the majority of AI research employing variations of Stochastic Gradient Descent (SGD) as the method from which a particular model learns. This trend, however, is not universal in nature, with some domains, such as time series analysis, persisting with use of alternative classification methods. Recent examples include Bai \textit{et al.}'s work applying ensemble learning\cite{BAI2021114162} and Yan \textit{et al.}'s work on time series similarity measurement \cite{yan2020extracting}. Additionally, research into efficient, alternative optimization strategies is also an active area of publication. Recent metaheuristic algorithms, such as the I-PKL-CS algorithm\cite{li2020improved}, the Dynamic Learning Evolution algorithm\cite{li2021dlea}, Elephant Herding Optimization\cite{li2020learning} (EHO), the Opposition-based Krill Herd algorithm\cite{wang2016opposition}, and EHO using dynamic topology and biogeography-based optimization\cite{li2021elephant}, have demonstrated efficient capabilities when optimizing towards a given solution, as explored by Li \textit{et al.}\cite{li2021survey} in a recent survey. \cite{li2020improved}\cite{li2021survey}\cite{li2021dlea}

Recent research employing SGD has enabled highly accurate models in certain sub-fields, such as computer vision, through use of Backpropagation-enabled Convolutional Neural Networks (CNNs). The first successful application came with AlexNet's \cite{krizhevsky2017imagenet} breakthrough performance on the ImageNet Large Scale Visual Recognition Competition in 2012. Since then, numerous CNN architectures have been proposed, with notable contributions to the field including VGGNet \cite{simonyan2014very}, ResNet \cite{he2016deep}, Inception V3 \cite{szegedy2016rethinking}, and DBN for image processing\cite{ying2018accelerating}. Medical image-based classification has specifically benefited from more performant computer vision techniques. Success has been found with use of SGD-based CNNs on a range of image-based medical domains. In recent literature, for example, Zhang \textit{et al.}\cite{zhang2021adoption} propose improved diagnosis of atrophic gastritis through application of DenseNet\cite{iandola2014densenet} and Wang \textit{et al.} propose MCNet\cite{wang2020multi} for use in automated lesion segmentation using endoscopy images of the gastrointestinal tract.

Achieving superhuman performance with today's models, however, comes with a steep requirement for data. The ILSVRC ImageNet dataset\cite{russakovsky2015imagenet}, for instance, contains over 14,000,000 images with roughly 21,000 image classes representing everyday things or objects. This dataset size is required in order to enable a performant understanding of each class. Additionally, the quality of a given dataset can be an issue, with class bias, class balance, and data quality all potential performance detractors, if neglected during model training. Even if these dataset considerations are put aside, modern approaches to AI can typically take multiple days to train on a challenging dataset. These limitations become particularly stifling when additional classes are considered for use with a model. Adding a new class to ImageNet requires roughly 600 new images to prevent class imbalance within the dataset. For scenarios involving common objects, such as those in ImageNet, obtaining 600 new images can be a fairly simple process. This changes, however, if images of the new class are difficult to obtain or existing datasets are incredibly shallow, resulting in class imbalance.

In an effort to solve the aforementioned issues, research into creating adaptable models for use on sparse datasets has seen active development in recent years. These efforts can generally be categorized into transfer learning and few-shot learning. This study focuses on the use of few-shot learning and its application to human cell classification.

The general goal of few-shot learning involves accurately performing a task on new data, given only a sparse amount of training data. Work in this field using CNNs largely began with Koch \textit{et al.} proposed Siamese Network\cite{koch2015siamese}, which demonstrated understanding of a new class given only a single "shot" or ground truth image. Vinyals \textit{et al.} furthered the field a year after with their proposed Matching Network\cite{vinyals2016matching} and additionally contributed the mini-ImageNet dataset, which is used for testing few-shot learning techniques today. Both of the previously mentioned approaches, however, encouraged quick recognition of new data through processes external to the model itself. Finn \textit{et al.} proposed Model-Agnostic Meta-Learning\cite{finn2017model} (MAML) instead approaches few-shot learning as a process by which a model's weights are directly manipulated. MAML's optimization-based approach enabled a significant jump in few-shot learning performance, with test accuracy improving by over 6\% on mini-Imagenet when compared to Vinyals et al.'s Matching Network. Building off of MAML's success, Nichol \textit{et al.} proposed Reptile \cite{nichol2018first}, a similar, optimization-based technique, which used a refined weight update strategy to achieve a 2\% accuracy increase over MAML on mini-ImageNet.

In the past 4 years since MAML, state-of-the-art few-shot learning techniques have shifted towards application of semi-supervised learning and transductive strategies for better performance. Current networks have improved significantly on mini-ImageNet, with the recently proposed Simple CNAPS + FETI\cite{Bateni_2020_CVPR} demonstrating 90.3\%, an improvement of over 27.2\% over MAML. Taking into consideration the recent progress within the field of few-shot learning, this study investigates whether few-shot learning techniques can be effectively applied to human cell classification in situations involving sparse datasets.

\begin{figure}[ht]
    \centering
    \includegraphics[width=0.9\textwidth]{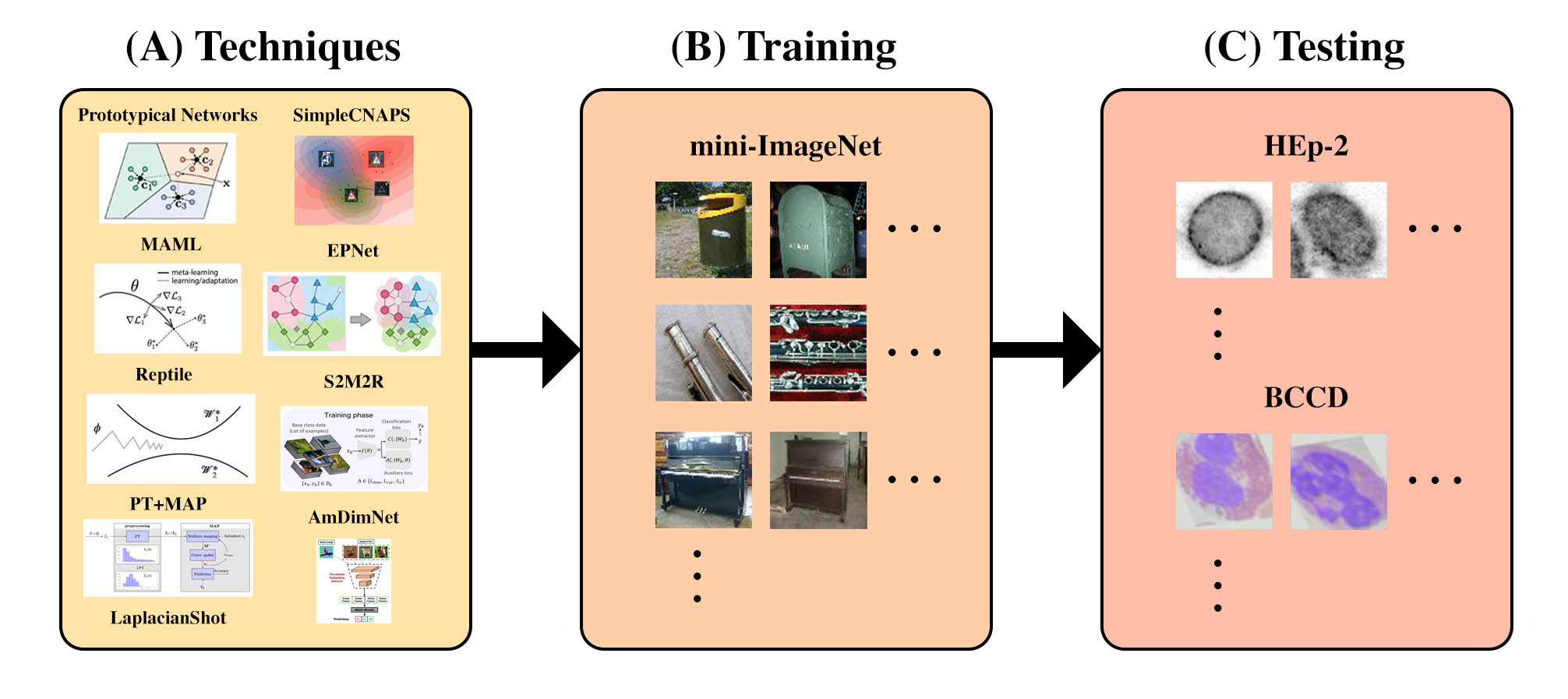}
    \caption{The process proposed for training and testing the nine selected few-shot learning techniques on out-of-domain data.}
    \label{fig:process}
\end{figure}

To summarize, the three main contributions are as follows:

\begin{enumerate}
\item This study investigates the use of few-shot learning in human cell classification. Figure \ref{fig:process} provides an illustrated example of the proposed process. To the best of the author's knowledge, this study is the first of its kind and will provide valuable insights to researchers in this field.
\item This study evaluates the potential benefits, if any, to varying the backbone architecture and training schemes in current state-of-the-art few-shot learning techniques when used in human cell classification.
\item This study presents future direction for research in this area based upon the findings from this study.
\end{enumerate}

The remainder of this paper is organized as follows: the methods sections details the few-shot learning techniques applied and the experimental regiment by which we apply them. The results section presents results from the aforementioned experiments. The discussion section explores our findings in further detail. The conclusions and future work section details the conclusions drawn from this study and establishes direction for future work performed in this area.

\section*{Methods}

\begin{figure}[ht]
    \centering
    \includegraphics[width=0.9\textwidth]{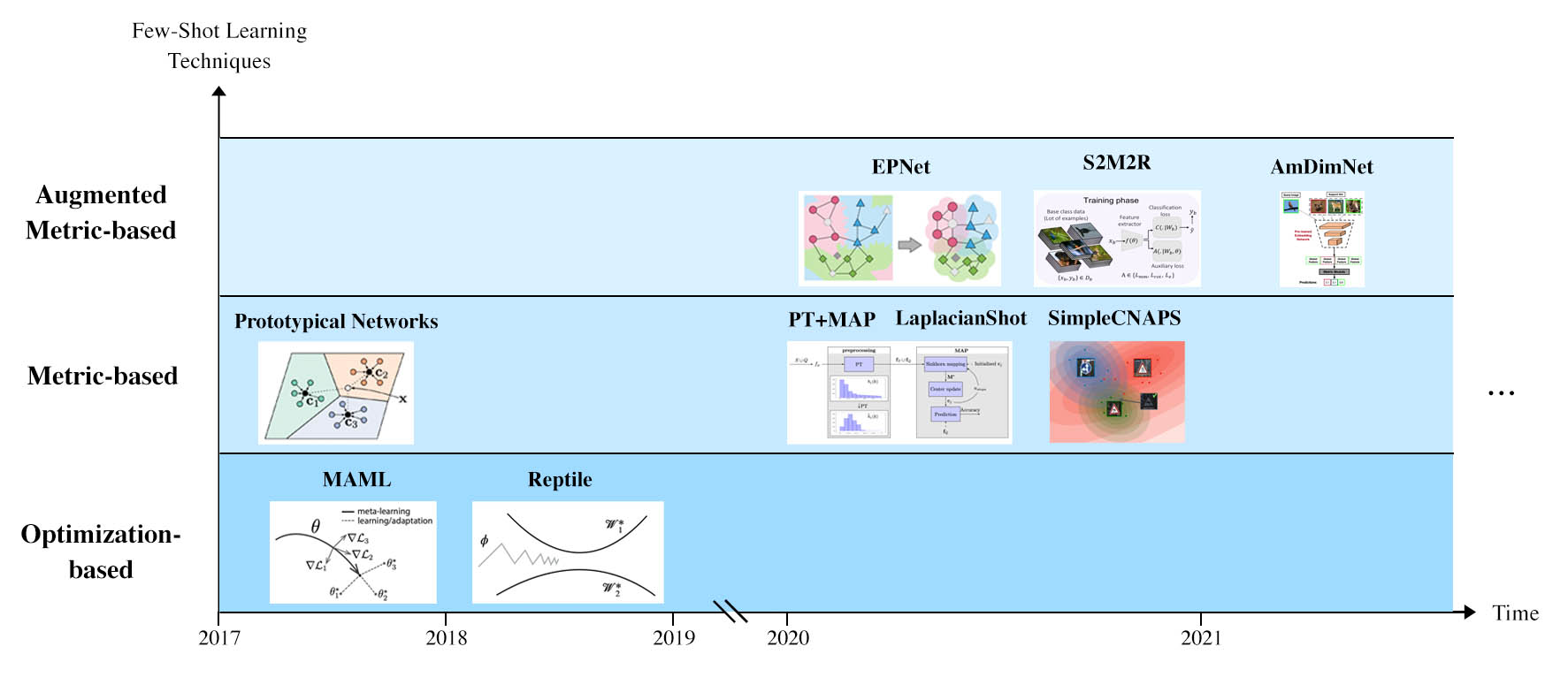}
    \caption{A temporal overview of notable few-shot learning techniques proposed within the past 5 years. "Optimization-based" few-shot learning techniques refer to those proposing changes to optimization processes employed by a network. "Metric-based" few-shot learning techniques refer to those proposing a metric from which a similarity score between a set of samples can be obtained from. "Augmented Metric-based" few-shot learning techniques refer to those proposing an augmentation (such as application of a self-supervised or transductive process) to a metric-based few-shot learning technique.}
    \label{fig:techniques}
\end{figure}

In the first part of this study, we train nine few-shot learning techniques on mini-ImageNet and evaluate their performance on two selected human cell datasets. This experimental setup allows for the model to train on a non-medical, balanced dataset and test few-shot performance on sparse medical datasets. The techniques used in this study were selected as the top nine from a set of notable, state-of-the-art techniques with code publicly available. Figure \ref{fig:techniques} provides an illustrated overview of the techniques investigated over time.

The second part of this study evaluates the potential benefits, if any, to varying the backbone architecture and training schemes in current state-of-the-art few-shot learning techniques. EPNet was selected as the experimental model due to its competitive few-shot learning performance on both medical datasets and efficient implementation.

All training and testing in this paper was performed using an NVIDIA Tesla V100 (with 32 GB of VRAM), PyTorch v1.8, and Python 3.8.

\subsection{Part 1: Investigating Existing Few-Shot Learning Techniques}

\subsubsection{Metric-based Few-Shot Learning}

Some of the earliest work within the field of few-shot learning leveraged metric-based analysis in order to generate a similarity score between two given samples. At a very general level, application of K-Nearest Neighbors to a given dataset can be a thought of as a rudimentary, metric-based few-shot learning model. Metric-based models consider input data similarly, clustering unlabelled data (known as the query set) based on information from a previously seen set of labelled data (known as the support set).

Snell \textit{et al.}'s Prototypical Networks\cite{snell2017prototypical} serve as notable, performant example of a recent metric-based few-shot learning approach. The methodology proposed establishes use of an embedding function to map a given query set and support set to an embedding space. The mean of each class within the support is taken and defined as a prototype vector. The squared euclidean distance between a query embedding and all prototype vectors is used to generate the final distribution over classes for a given query point.

\subsubsection{Optimization-based Few-Shot Learning}

In contrast to metric-based strategies, approaches leveraging optimization-based few-shot learning propose no external metrics by which the model depends on. Instead, a model-agnostic approach is taken by defining a general-purpose optimization algorithm compatible with all models leveraging Stochastic Gradient Descent-based methods for learning. By applying this algorithm, all potential classes are optimized, rather than continuous optimization towards a single dataset.

To enable further exploration and understanding of this strategy, we define a generic model as $\textit{f}_{\theta}$ with parameters $\theta$, a generic dataset, $\textit{D}$, a learning rate $\alpha$, and a generic loss function, $\textit{L}$. A "task" $\textit{T}_{i}$ is sampled from a dataset $\textit{D}$ as a grouping of classes. With the defined variables, we can update by a single Stochastic Gradient Descent iteration using the following equation:

\begin{align}
\label{eq:1}
\theta_{i}^{'} = \theta - \alpha\nabla_{\theta}\textit{L}_{\textit{T}_{i}}(\textit{f}_{\theta})
\end{align}

\begin{figure}[ht]
    \centering
    \includegraphics[width=0.25\textwidth]{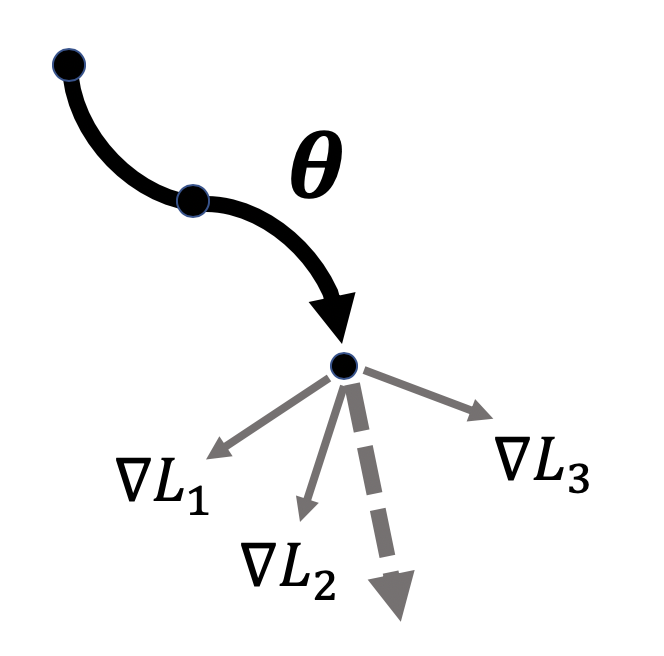}
    \caption{The optimization-based process for optimizing towards three tasks illustrated.}
    \label{fig:maml}
\end{figure}

In doing so, however, we only compute the loss for a single batch within a single task. Optimization-based approaches, such as MAML and Reptile, promote accuracy across all given tasks, rather than a single task. To achieve generalization towards new tasks, MAML (and additionally Reptile) proposes an adaptation process which involves taking multiple gradient descent iterations for each task. The loss of each task is analyzed, enabling discovery of the optimal $\theta^{*}$ which optimizes towards all tasks. In essence, Equation (\ref{eq:1}) is used to take small, iterative gradient descent steps for each task, discovering how optimization occurs. Once all tasks have been iterated over, $\theta^{*}$ can be found, enabling us to take a large step in an overall optimal direction. Figure \ref{fig:maml} illustrates optimizing across three given tasks. With the above equation, we would have only taken a single step along either $\nabla\textit{L}_{1}$, $\nabla\textit{L}_{2}$, or $\nabla\textit{L}_{3}$. 

\subsubsection{Transductive and Self-Supervised Approaches to Few-Shot Learning}

\begin{figure}[ht]
    \centering
    \includegraphics[width=0.5\textwidth]{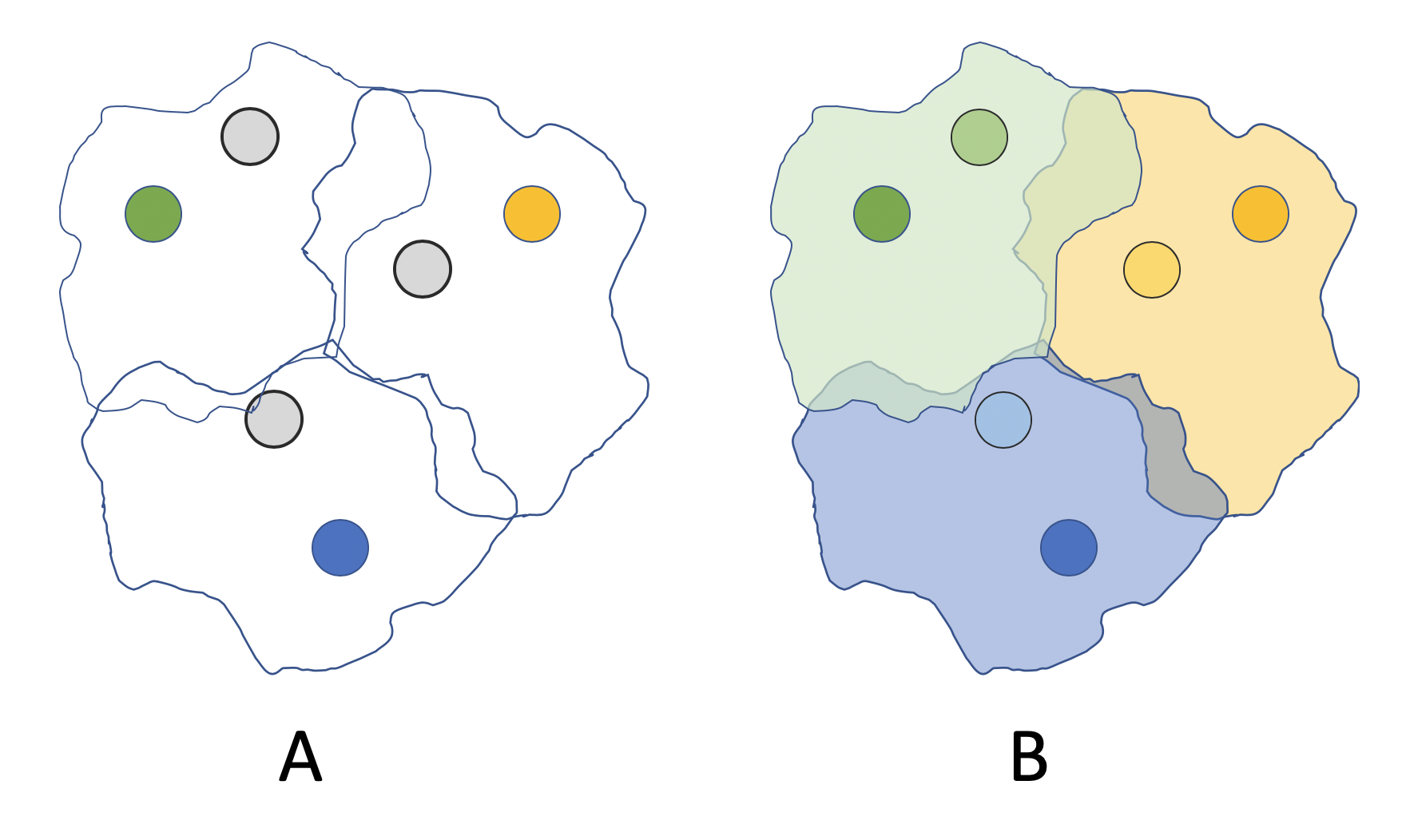}
    \caption{An illustrated example of transductive few-shot learning. (A) Grey circles represent unlabelled points (the query set) and coloured circles represent labelled points (the support set). (B) All unlabelled points are labelled based on their position within the labelled data.}
    \label{fig:epnet}
\end{figure}

Recent state-of-the-art studies within the field of few-shot learning have demonstrated use of transductive techniques, self-supervised learning, and extra, unlabelled data in order to enable accurate performance. Rodriguez \textit{et al's} EPNet\cite{rodriguez2020embedding} follows a transductive few-shot learning approach to enable quick uptake of new classes. In contrast, where optimization-based approaches leveraged a modified gradient descent algorithm, EPNet maps the support and query sets to an embedding space wherein all points are considered simultaneously. During this phase, labels are propagated from the support set to similar, unlabelled query set points. Figure \ref{fig:epnet} illustrates the process of propagation for a given set of points. Bateni \textit{et al.}'s proposed Simple CNAPS\cite{Bateni_2020_CVPR} follows a similar metric-based clustering, however, a Mahalanobis distance is used for comparison between points, rather than propagation of labels. PT+MAP\cite{hu2020leveraging} and LaplacianShot\cite{ziko2020laplacian} function similarly, however, both propose alternative strategies for distance metrics when considering query and support points. AmdimNet\cite{chen2021self} and S2M2\cite{mangla2020charting}, alternatively, leverage self-supervised techniques in order to generate a stronger embedding-space mapping for input data. 

\subsubsection{Dataset Selection and Few-Shot Data Sampling}

All experiments performed in this study used the mini-ImageNet few-shot dataset\cite{vinyals2016matching} for training purposes. The BCCD White Blood Cell Dataset\cite{BCCD2019} and a contrast-adjusted variant of the HEp-2 Cell dataset\cite{larsen2014hep} were used for out-of-domain few-shot testing. The HEp-2 dataset was additionally used for training. The aforementioned datasets were sampled from following the procedure defined by Vinyals et al.\cite{vinyals2016matching} for few-shot datasets.

Sampling data from the training and testing few-shot datasets was performed in an identical manner. Batches sampled from a dataset are defined as "episodes" for a given model. These episodes are composed of two data sections: a support set and a query set. The support set is comprised of labelled examples meant to teach a given model about the episode's classes. Labelled classes, specifically, are sampled in terms of ways and shots. The number of ways in an episode defines how many classes are sampled, while the number of shots defines how many of each class is sampled. The query set, in contrast, is composed of unlabelled data and serves as what can be considered a miniature "test set" for the episode to benchmark how well a given model learned from the support set. Typically, 10 images per way are sampled from the respective class. All experiments performed in this study use 5-way 5-shot sampling strategies for testing and training. Additionally, due to the sampling-based few-shot datasets lacking a clear end, we define one epoch as being equivalent to sampling 600 episodes from a given dataset.

\subsection{Part 2: Evaluating Modifications}

\subsubsection{Data Augmentations and Regularization}

We apply data augmentation and regularization techniques used in recent classification approaches in an attempt to prevent overfitting on mini-ImageNet and encourage a more general embedding space. Hyperparameters associated with the employed techniques applied in two ways. First, the hyperparameter is set to a static value and the model is left to train. Second, we follow the approach detailed in the recent EfficientNetV2\cite{tan2021efficientnetv2} training scheme and gradually increase (or decrease) a hyperparameter's value over a number of epochs.

\paragraph{RandAugment\cite{cubuk2020randaugment}} A series of N random data augmentations (shear, colour jitter, etc) are applied to a batch of images at a magnitude of M. N was set at a static value of 2 (an optimal value defined by Cubuk \textit{et al.}) for all experiments and M was tested at ranges [5-25], [5-15], [5-10], and a static value of 5.

\paragraph{Mixup\cite{zhang2017mixup}} During the creation of an episode, sampled images are blended with another random image. The blending amount for one image is defined by the alpha hyperparameter. The blending amount for the alternate image is defined as $1 - alpha$. Alpha was tested at range [0-0.25], [0-0.1], and at a static value of 0.1.

\paragraph{Label Smoothing\cite{muller2019does}} One-hot encoded vectors output from a classification model are run through label smoothing before the final loss calculation is performed. In doing so, the one-hot encoded vector is replaced with a smoothed, uniform distribution over the number of classes in the vector. Label smoothing contains a single hyperparameter that controls the level of smoothing applied to a given one-hot vector. The hyperparameter was tested at a static value of 0.1 (as defined by Muller \textit{et al.}).

\paragraph{Exponential Moving Average of Model Weights\cite{kingma2014adam}} An exponential moving average is retained of a given model's weights while training is underway. Averaged weights have been shown to perform better in some scenarios when applied to classification-based tasks.

\subsubsection{Architectural Improvements}

Internal changes to a given model were investigated during experimentation. Hu et al.'s Squeeze and Excitation\cite{hu2018squeeze} was evaluated due to its flexible implementation and notable performance improvement. This architectural improvement, however, was noted to work best with Residual Networks and, as such, we only evaluate this architectural change using the ResNet-12 backbone. A hyperparameter R is exposed with use of Squeeze and Excitation, enabling variation in the computational cost of the Squeeze and Excitation block it pertains to. We tested this value at 0.1, 0.25, and 2.0.

\subsubsection{Training Methods}

We adopt two training strategies during experimentation. The first, and more prevalent, strategy involves a single, long training cycle, typically defined as 100 epochs long on mini-ImageNet. At the end of this cycle, we expect the model's loss to have converged. A plateau-based learning rate scheduler is used to manage the learning rate throughout the training process. The second training strategy we employ, involves mirroring the EfficientV2 training setup as part of the few-shot training process. Instead of a single, long cycle, we substitute with multiple, shorter training cycles (all training the same model), which we define as 80 epochs long. Four cycles are performed in total, which results in 320 epochs of training. Over the course of each consecutive cycle, the model is exposed to images that gradually become larger, all the while regularization is increased to prevent overfitting. An exponential decay learning rate schedule is used over each cycle, ensuring consistent performance.

\subsubsection{Network Backbone Evaluation}

Within the field of few-shot learning, the majority of proposed methodologies have historically leveraged one of three network backbones: a 4CONV network proposed as part of Vinyals' Matching Network\cite{vinyals2016matching}, a Wide Residual Network\cite{zagoruyko2016wide} with depth 28 and width 10, or a Residual Network\cite{he2016deep} with 12 layers. To investigate application of other network backbones, we employ use of three networks: EfficientNetV2-Small\cite{tan2021efficientnetv2}, ResNet-18, and DenseNet\cite{lichtenstein2020tafssl}. EfficientNetV2 was selected for its state-of-the-art performance on modern, image classification datasets. ResNet-18 and DenseNet were selected for their state-of-the-art performance in previously proposed few-shot learning techniques.

\section*{Results}

\subsection{Part 1: Investigating Existing Few-Shot Learning Techniques}

\subsubsection{Model Evaluation}

To aid in comparing each model, Table \ref{table:4} provides a detailed overview of the respective technical attributes. The column headings within the table are expanded upon as follows.

\paragraph{Technique}
The type or style of few-shot learning applied.

\paragraph{Backbone}
The backbone network applied by the few-shot learning approach. WRN28-10 refers to the WideResNet28-10 model. CONV4 refers to the convolutional model proposed by Vinyals \textit{et al.} in MAML.

\paragraph{Preprocessing}
Whether or not input to the network requires preprocessing by a feature extractor network.

\paragraph{Extra Training Data}
Whether or not extra unlabelled training data was used to boost performance. Unlabelled training data typically relates to the support set at hand (for example, additional unlabelled images of cats are used if the cat class is in the support set).

\begin{table}
\centering
\caption{An overview of the differing details between the models trained and tested.}
\begin{tabular}{ |p{3cm}|p{4cm}|p{2cm}|p{2cm}|p{3cm}| }
 \hline
 \multicolumn{5}{|c|}{Model Evaluation Table} \\
 \hline
 Model Name & Technique & Backbone & Preprocessing & Extra Training Data \\
 \hline
 AmdimNet\cite{chen2021self} & Self-supervised Metric & AmdimNet & No & Yes \\
 \hline
 EPNet\cite{rodriguez2020embedding} & Transductive Metric & WRN28-10 & No & Yes \\
 \hline
 SimpleCNAPS\cite{Bateni_2020_CVPR} & Metric & ResNet18 & No & Yes \\
 \hline
 PT+MAP\cite{hu2020leveraging} & Metric & WRN28-10 & Yes & No \\
 \hline
 LaplacianShot\cite{ziko2020laplacian} & Metric & WRN28-10 & No & No \\
 \hline
 S2M2R\cite{mangla2020charting} &  Self-supervised Metric & WRN28-10 & Yes & No \\
 \hline
 Reptile\cite{nichol2018first} & Optimization & CONV4 & No & No \\
 \hline
 MAML\cite{finn2017model} & Optimization & CONV4 & No & No \\
 \hline
 ProtoNet\cite{snell2017prototypical} & Metric & CONV4 & No & No \\
 \hline
\end{tabular}
\label{table:4}
\end{table}

\subsubsection{Baseline Benchmarks}

Nine few-shot learning techniques were trained and benchmarked using the mini-ImageNet training, validation, and testing dataset splits. Training was performed for 100 epochs using a plateau-based learning rate scheduler with a patience of 10. From our observations, 100 epochs were sufficient to reach convergence in all model permutations. Convergence was typically reached around the 60-80 epoch. Other training settings (optimizer choice, hyperparameter values, etc) were reproduced as defined by author's of each respective technique. A summary of these settings can be found in Table \ref{table:details}. The test accuracy on mini-ImageNet reported for all models was reproduced within margin-of-error. Out-of-domain few-shot testing was performed using BCCD and HEp-2 as a 5-way 5-shot experiment. Performance demonstrated by all models on the HEp-2 dataset was within expectations, however,  the accuracy exhibited after testing on BCCD suggested potential problems. The issues were identified as problems stemming from input image size. Mini-ImageNet images have a resolution of 84px by 84px while BCCD images have a resolution of 224px by 224px. During initial testing experiments, all input images were resized to 84px by 84px through anti-aliased, local mean downsampling. The featured white blood cell in a BCCD image that is being classified is not typically a prominent feature within an input image, thus, by resizing, a significant portion of detail was lost, degrading few-shot performance. Global pooling layers were added to all models, enabling arbitrary input image size and significantly improving BCCD few-shot performance. Table \ref{table:1} contains all results from the baseline tests run.

\begin{table}
\centering
\caption{Parameter details specific to each technique.}
\begin{tabular}{ p{3cm}| p{2cm} p{2cm} p{2cm} p{2cm}  }
 \hline
 \multicolumn{5}{|c|}{Technique Implementation Details} \\
 \hline
 Model & Optimizer & Momentum & Weight Decay & Batch Size\\
 \hline
 AmdimNet\cite{chen2021self} & Adam & - & - & 100 \\
 EPNet\cite{rodriguez2020embedding} & SGD & 0.9 & 0.0005 & 128 \\
 SimpleCNAPS\cite{Bateni_2020_CVPR} & Adam & - & - & 256 \\
 PT+MAP\cite{hu2020leveraging} & Adam & - & - & 16 \\
 LaplacianShot\cite{ziko2020laplacian} & SGD & 0.9 & 0.0001 & 128 \\
 S2M2R\cite{mangla2020charting} & Adam & - & - & 16 \\
 Reptile\cite{nichol2018first} & Adam & - & - & 5 \\
 MAML\cite{finn2017model} & Adam & - & - & 32 \\
 ProtoNet\cite{snell2017prototypical} & Adam & - & - & 5 \\
 \hline
\end{tabular}
\label{table:details}
\end{table}

\begin{table}
\centering
\caption{Test accuracy results from baseline experiments run against the mini-ImageNet test set, BCCD, and HEp-2. Testing using the BCCD dataset was performed using additional global pooling layers.}
\begin{tabular}{ p{3cm}| p{3cm} p{3cm} p{3cm}   }
 \hline
 \multicolumn{4}{|c|}{Initial Dataset Test Performance} \\
 \hline
 Model & mini-ImageNet & BCCD & HEp-2\\
 \hline
 AmdimNet\cite{chen2021self} & 89.75 $\pm$ 0.12 & 48.35 $\pm$ 0.18 & 54.32 $\pm$ 0.21 \\
 EPNet\cite{rodriguez2020embedding} & 88.66 $\pm$ 0.24 & 47.39 $\pm$ 0.22 & \textbf{55.12 $\pm$ 0.13} \\
 SimpleCNAPS\cite{Bateni_2020_CVPR} & \textbf{90.11 $\pm$ 0.17} & 47.06 $\pm$ 0.72 & 53.15 $\pm$ 0.84 \\
 PT+MAP\cite{hu2020leveraging} & 88.02 $\pm$ 0.13 & 42.94 $\pm$ 0.17 & 54.73 $\pm$ 0.22 \\
 LaplacianShot\cite{ziko2020laplacian} & 82.27 $\pm$ 0.15 & 34.75 $\pm$ 0.13 & 44.69 $\pm$ 0.17 \\
 S2M2R\cite{mangla2020charting} & 82.81 $\pm$ 0.31 & 44.15 $\pm$ 0.23 & 54.41 $\pm$ 0.27 \\
 Reptile\cite{nichol2018first} & 65.62 $\pm$ 0.28 & \textbf{50.91 $\pm$ 0.12} & 51.76 $\pm$ 0.13 \\
 MAML\cite{finn2017model} & 64.62 $\pm$ 0.19 & 42.81 $\pm$ 0.21 & 45.21 $\pm$ 0.24 \\
 ProtoNet\cite{snell2017prototypical} & 67.88 $\pm$ 0.12 & 46.89 $\pm$ 0.13 & 50.70 $\pm$ 0.17 \\
 \hline
\end{tabular}
\label{table:1}
\end{table}

In an effort to further explore cell image-based few-shot performance, in-domain training and testing was performed using HEp-2 as the training dataset and BCCD as the testing dataset. HEp-2 was selected as the training dataset due to the larger number of classes present (6 classes) versus BCCD (5 classes). In-domain training and testing was performed in the same manner as out-of-domain testing. The top performing techniques from out-of-domain testing (Reptile on BCCD and EPNet on HEp-2) were used. Table \ref{table:in-domain} details the results obtained from the in-domain tests run.

\begin{table}
\centering
\caption{Test accuracy results from in-domain training on HEp-2 and testing on BCCD.}
\begin{tabular}{ p{3cm}| p{3cm} }
 \hline
 \multicolumn{2}{|c|}{In-domain Performance} \\
 \hline
 Model & BCCD \\
 \hline
 EPNet\cite{rodriguez2020embedding} & 45.31 $\pm$ 0.21 \\
 Reptile\cite{nichol2018first} & 40.24 $\pm$ 0.23 \\
 \hline
\end{tabular}
\label{table:in-domain}
\end{table}

\subsection{Part 2: Evaluating Modifications}

\subsubsection{Backbone Variations}

Three differing styles of network backbone were evaluated in an attempt to further increase few-shot performance on EPNet. We solely train and test on mini-ImageNet in this instance since high accuracy on mini-ImageNet results in high accuracy on out-of-domain datasets. To evaluate each backbone, EPNet's original WideResNet backbone was replaced, trained, and tested with EfficientNetV2, ResNet-18, and DenseNet. All selected backbone replacements, however, failed to match or surpass the original WideResNet backbone. This result could likely be due to the relative complexity some of the selected backbones exhibited. Table \ref{table:2} contains a detailed breakdown of the experimental results. DenseNet had already demonstrated application in a recent few-shot learning approach, thus, the closest result being attributed to this network is no surprise.

\begin{table}
\centering
\caption{Test accuracy results from using different backbone variations in EPNet on and testing on mini-ImageNet, BCCD, and HEp-2. Each backbone was trained on mini-ImageNet's training set before testing.}
\begin{tabular}{ p{6cm}| p{3cm} p{2cm} p{2cm} }
 \hline
 \multicolumn{4}{|c|}{Backbone Performance} \\
 \hline
 Backbone & mini-ImageNet & BCCD & HEp-2\\
 \hline
 \textbf{WideResNet28-10 (Original Backbone)} & \textbf{88.7\%} & \textbf{47.4\%} & \textbf{55.1\%} \\
 EfficientNetV2 (Default Width) &  59.8\% & 18.3\% & 26.3\% \\
 EfficientNetV2 (0.5 Width) &  67.3\% & 25.7\% & 33.3\% \\
 EfficientNetV2 (0.75 Width) &  69.2\% & 28.0\% & 35.7\% \\
 EfficientNetV2 (2.75 Width) &  70.8\% & 29.5\% & 37.1\% \\
 ResNet-18 & 68.2\% & 26.8\% & 34.7\% \\
 DenseNet & 78.8\% & 37.4\% & 45.0\% \\
 \hline
\end{tabular}
\label{table:2}
\end{table}

\subsubsection{Model Additions}

Various model additions were added to EPNet and benchmarked using mini-ImageNet. All additions were trained for 100 epochs with a plateau-based learning rate schedule. Table \ref{table:3} contains the full list of addition evaluation results. Generally, all proposed additions had a negative impact on EPNet during training. Some additions decreased accuracy by a couple percent while others drained accuracy by a large amount.

\begin{table}
\centering
\caption{Test accuracy results using different model additions within EPNet. Each model addition was independently trained on the mini-ImageNet training set and tested on the mini-ImageNet test set, BCCD, and HEp-2}
\begin{tabular}{ p{6cm}|p{3cm} p{2cm} p{2cm} }
 \hline
 \multicolumn{4}{|c|}{Model Addition Performance} \\
 \hline
 Model Addition & mini-ImageNet & BCCD & HEp-2\\
 \hline
 \textbf{No Additions} & \textbf{88.7\%} & \textbf{47.4\%} & \textbf{55.12\%} \\
 RandAugment (Magnitude=5-15) &  75.8\% & 34.6\% & 42.1\% \\
 RandAugment (Magnitude=5-10) &  69.3\% & 27.8\% & 35.6\% \\
 RandAugment (Magnitude=5) &  70.1\% & 28.9\% & 36.1\% \\
 Squeeze and Excitation (Reduction=0.10) & 68.7\% & 27.6\% & 35.2 \\
 Squeeze and Excitation (Reduction=0.25) & 68.2\% & 26.2\% & 34.9\% \\
 Squeeze and Excitation (Reduction=2) & 63.9\% & 22.4\% & 30.5\% \\
 Mixup (Alpha=0.10) & 76.2\% & 34.7\% & 42.8\% \\
 Label Smoothing (A=0.10) & 65.3 & 23.2\% & 31.4\% \\
 Exponential Moving Average & 78.6\% & 38.2\% & 44.1\% \\
 \hline
\end{tabular}
\label{table:3}
\end{table}

\section*{Discussion}

Analyzing and classifying human cells (such as in blood smears or skin biopsies) is an intensive task requiring specialized equipment and oversight from a trained professional. With recent progress in computer vision performance, however, automated image-based analysis of human cells has been an active area of research. Modern deep learning-based approaches have specifically enabled superhuman performance in a wide array of fields. Application of deep learning to medical scenarios, however, has typically stagnated due to dataset size requirements. A potential solution to these issues lies within the field of few-shot learning, an area of research concerned with building performant networks using sparse amounts of data. Recent few-shot learning-based approaches have demonstrated increasingly accurate performance on complex dataset, such as mini-ImageNet. In this study, we investigated whether few-shot learning-based techniques could mitigate the data requirements necessary for performant deep learning-based cell classification. An optimal scenario, in this regard, would involve a selected few-shot approach training on a non-medical dataset and accurately testing on a sparse medical dataset. Successful application of a few-shot technique to sparse medical data would drastically expedite existing workflows, potentially allowing automation of tasks typically allocated to trained professionals.

To facilitate this study, we selected mini-ImageNet\cite{vinyals2016matching}, a popular benchmark for few-shot learning techniques, as the dataset by which we would train on. For human cell-based evaluation, we selected the BCCD Dataset\cite{BCCD2019} (BCCD) and the HEp-2 Dataset\cite{larsen2014hep} as the testing datasets. Our experimental process involved training few-shot approaches on mini-ImageNet and testing the resulting models on the BCCD dataset and the HEp-2 dataset. In doing so, we benchmarked embeddings learned from a non-medical dataset on human cell-based classification. We selected 9 notable, few-shot learning models proposed over the past 5 for use in our experiment. Each model was implemented using the authors' code (if available) and trained from scratch. Before testing, each model's performance on mini-ImageNet was verified against the original reported results (within margin-of-error).

After completing experimentation, a decrease in accuracy of at least 30\% was noted when transitioning from the training dataset to an out-of-domain human cell dataset. In a rather shocking result, however, Reptile, a relatively old technique, out-performed all newer few-shot learning approaches on the BCCD dataset and performed competitively on the HEp-2 dataset. MAML, a similar technique, also performed competitively on the out-of-domain testing datasets, beating a few newer approaches as well. These results potentially indicate that relatively high performance on mini-ImageNet (and other few-shot benchmarks) does not necessarily guarantee proportional performance on out-of-domain tests. Reptile and MAML's optimization-based strategy for fast adaption to new classes could also lead to further performance in out-of-domain tests.

Overall, performance degradation on the selected medical datasets can largely be attributed to difficulties transitioning from a non-medical domain to a medical domain. Severe out-of-domain accuracy decreases in few-shot learning are corroborated by Bateni \textit{et al.} in their experimentation with SimpleCNAPS\cite{Bateni_2020_CVPR}. Decreases as large as 20\% were noted for out-of-domain images within the same dataset. This decrease, however, is a significant issue when rigorous standards for medical practice are taken into account. Models aiming for deployment in medicinal scenarios typically demonstrate high accuracy in their field of application. Even with proven and accurate capabilities, results produced by a model in a medical setting are still rigorously reviewed. Introducing one of the current few-shot learning approaches investigated in this study could potentially lead to incorrect output or, at worst, misdiagnosis for a patient.

In an attempt to boost few-shot performance, a variety of architectural revisions, data augmentation approaches, and training schemes were experimented with and benchmarked using mini-ImageNet. EPNet was selected for this experimentation due to its ease of implementation and performance. Through this process, we discovered that recent measures taken to improve classification networks are ineffective on few-shot learning-based networks. For example, EPNet's backbone network was swapped and trained across a selection of performant, state-of-the-art classification backbones. EfficientNetV2\cite{tan2021efficientnetv2}, a network that recently achieved state-of-the-art accuracy on ImageNet, results in an accuracy decrease of at least 12\% when combined with EPNet. Similar, performance-boosting classification techniques, such as data regularization, resulted in performance regressions. After an exhaustive exploration of the aforementioned techniques, we concluded that the employed few-shot learning technique should be the main focus for performance-based changes.

The methods applied in this work largely focus on a single training dataset and two, cell-based out-of-domain testing datasets. Other, more rigorous few-shot learning evaluation frameworks, such as Triantafillou \textit{et al.}'s Meta-Dataset\cite{triantafillou2019meta}, employ use of multi-dataset strategies to gain a clearer understanding of a model's performance. Furthermore, the few-shot learning techniques used in this study are selected within a limited window of time (5 years).

\section*{Conclusions and Future Work}

In this study, we investigate the use of few-shot learning in human cell classification. During the performed training and testing, a variety of backbone architectures and training schemes were benchmarked for any potential benefit. Although all tested techniques performed well when classifying unseen training data, significant performance decreases were observed when transitioning to either of the two human cell classification testing datasets. With this in mind, we believe that few-shot learning techniques are still limited in the scope of problems they can solve. Support for new techniques less "brittle to [the] narrow domains they were trained on" was recently highlighted by Turing award winners Bengio, LeCun, and Hinton \cite{10.1145/3448250}. As such, we posit that a stronger emphasis on out-of-domain robustness should be one of the main directions for future few-shot learning research.

In conclusion, few-shot learning methodologies are not yet capable of accurately performing out-of-domain classification at a level accurate enough for human cell identification. We test this conclusion across a selection of notable few-shot learning models proposed within the last 5 years. After training on mini-ImageNet and testing on the BCCD and HEp-2 datasets, performance was found to drop by at least 30\% after transitioning from the non-medical dataset to the selected medical datasets. With this in mind, application of current few-shot learning methodology to medical scenarios is, at this time, insufficient.

To facilitate better out-of-domain performance in few-shot learning, new areas of exploration are necessary. MAML and Reptile's surprisingly competitive out-of-domain performance underscores a need to reconsider older few-shot learning techniques. With this in mind, future few-shot learning research should reevaluate optimization strategies or focus on more flexible few-shot distance metrics. Metaheuristic algorithms, such as Monarch Butterfly Optimization\cite{wang2019monarch}, the Earthworm Optimization Algorithm\cite{wang2018earthworm}, Elephant Herding Optimization\cite{wang2015elephant}, the Moth Search algorithm\cite{wang2018moth}, the Slime Mould algorithm\cite{li2020slime}, and Harris Hawks optimization\cite{heidari2019harris}, serve as possible directions for improvement in optimization-based techniques. In future work, we plan on revisiting this area of research and investigating a wider variety of few-shot learning approaches across a more comprehensive set of datasets (in and out-of-domain).

\bibliography{main}

\section*{Acknowledgements}

Funding for this project was provided through the MITACS Accelerate grant.

\section*{Author contributions statement}

R.W, M.H.A, M.S.S, and M.M.M conceived the experiments, R.W conducted the experiments, R.W, M.H.A, M.S.S, and M.M.M analysed the results. All authors reviewed the manuscript.

\section*{Ethics Declarations}

\paragraph{Competing Interests}

The authors declare no competing interests.

\end{document}